\def\arxivlook{report}

\newif\ifLookReport
\newif\ifLookTwocol
\def\LookReportName{report}
\def\LookTwocolName{twocol}
\ifx\arxivlook\LookReportName \LookReporttrue \fi
\ifx\arxivlook\LookTwocolName \LookTwocoltrue \fi

\ifLookReport
\documentclass[11pt,letterpaper]{article}
\else\ifLookTwocol
\documentclass[10pt,letterpaper,twocolumn]{article}
\else
\documentclass[10pt,letterpaper]{article}
\fi\fi

\newcommand{\papertitle}{Experts Across Schemas: Agent-Guided Mixture-of-Relational-Experts for Cross-Database Generalization}
\newcommand{\shorttitle}{Experts Across Schemas}
\newcommand{\pdfauthors}{Hiren Madhu, Sukanta Ganguly, Kiran Srinivasan, Smita Krishnaswamy, Rex Ying}
\newcommand{\codeurl}{https://anonymous.4open.science/r/MoREGT-7BDC/README.md}

\newcommand{\authorname}[2]{\mbox{\textbf{#1}\textsuperscript{#2}}}
\newcommand{\authorsep}{\hskip 1em plus 0.4em minus 0.4em\relax}
\newcommand{\affiliations}[1]{\par\vspace{0.45em}{\small #1\par}}

\title{\papertitle}
\author{%
  \authorname{Hiren Madhu}{1}\thanks{Correspondence to \texttt{hiren.madhu@yale.edu}. Work done at NetApp.}\authorsep
  \authorname{Sukanta Ganguly}{2}\authorsep
  \authorname{Kiran Srinivasan}{2}\authorsep
  \authorname{Smita Krishnaswamy}{1}\authorsep
  \authorname{Rex Ying}{1}%
  \affiliations{\textsuperscript{1}Yale University, USA\qquad\textsuperscript{2}NetApp, USA}%
}
\date{\today}

\usepackage[T1]{fontenc}
\usepackage{amsmath}
\usepackage{amsthm}  
\ifLookReport
  \usepackage{XCharter}
  \usepackage[xcharter,vvarbb]{newtxmath}
  \usepackage{sourcesanspro}
  \usepackage[letterpaper,margin=1in,headheight=14pt]{geometry}
\else
  \usepackage{newtxtext}
  \usepackage{newtxmath}
  \ifLookTwocol
    \usepackage[letterpaper,textwidth=7in,textheight=9.2in,columnsep=0.25in,
                centering,headheight=14pt]{geometry}
  \else
    \usepackage[letterpaper,textwidth=5.5in,textheight=9in,top=1in,
                hcentering,headheight=14pt,footskip=30pt]{geometry}
  \fi
\fi
\usepackage[varqu,scaled=0.95]{zi4}
\usepackage{microtype}

\usepackage{graphicx}
\usepackage{booktabs}
\usepackage{array}
\usepackage{tabularx}
\usepackage{adjustbox}
\usepackage{enumitem}
\usepackage[table]{xcolor}
\usepackage{pifont}
\usepackage{titlesec}
\usepackage{fancyhdr}
\usepackage[font=small,labelfont=bf]{caption}
\ifLookTwocol
  \usepackage[numbers,square,sort&compress]{natbib}
\else
  \usepackage[round]{natbib}
\fi
\ifLookReport
  \usepackage[skins]{tcolorbox}
\fi

\definecolor{Accent}{HTML}{1F4E79}
\definecolor{LinkCol}{HTML}{1A5A96}
\usepackage[colorlinks=true,linkcolor=LinkCol,citecolor=LinkCol,urlcolor=LinkCol]{hyperref}
\hypersetup{pdftitle={\papertitle},pdfauthor={\pdfauthors}}
\usepackage{xurl}  
\newcommand{\pmstd}[1]{{\scriptsize$\pm$#1}}
\newcommand{\stepnum}[1]{\ding{\numexpr 181+#1\relax}}
\newcommand{\figw}[2]{\ifLookTwocol #2\else #1\fi}

\theoremstyle{definition}
\newtheorem{definition}{Definition}[section]

\makeatletter
\ifLookReport
  \titleformat{\section}{\sffamily\Large\bfseries\color{Accent}}{\thesection}{0.7em}{}
  \titleformat{\subsection}{\sffamily\large\bfseries}{\thesubsection}{0.7em}{}
  \titlespacing*{\section}{0pt}{3.0ex plus 1ex minus .2ex}{1.4ex plus .2ex}
  \titlespacing*{\subsection}{0pt}{2.4ex plus 1ex minus .2ex}{1.0ex plus .2ex}
  \renewcommand{\@maketitle}{%
    \begingroup\raggedright\parindent\z@
    {\sffamily\footnotesize\bfseries\color{Accent}\textls[120]{PREPRINT}\hfill
     \mdseries\color{black!60}\@date\par}
    \vskip 0.5em{\color{Accent}\hrule height 0.8pt}\vskip 1.6em
    {\sffamily\bfseries\fontsize{21}{26}\selectfont\@title\par}
    \vskip 1.5em
    {\raggedright\@author\par}
    \vskip 1.6em
    \endgroup}
  \newtcolorbox{AbstractBox}{enhanced,colback=Accent!5,colframe=Accent!5,
    boxrule=0pt,sharp corners,left=12pt,right=12pt,top=9pt,bottom=10pt,
    borderline west={2.5pt}{0pt}{Accent}}
  \renewenvironment{abstract}{%
    \begin{AbstractBox}\small
    {\sffamily\bfseries\color{Accent}Abstract}\par\vspace{0.45em}}%
    {\end{AbstractBox}\vspace{0.6em}}
  \fancypagestyle{plain}{\fancyhf{}%
    \fancyfoot[C]{\sffamily\footnotesize\color{black!55}\thepage}}
\else\ifLookTwocol
  \titleformat{\section}{\large\bfseries}{\thesection}{0.8em}{}
  \titleformat{\subsection}{\normalsize\bfseries}{\thesubsection}{0.8em}{}
  \titlespacing*{\section}{0pt}{2.2ex plus .8ex minus .2ex}{1.0ex plus .2ex}
  \titlespacing*{\subsection}{0pt}{1.8ex plus .6ex minus .2ex}{0.8ex plus .2ex}
  \renewcommand{\@maketitle}{%
    \vbox{\hsize\textwidth\linewidth\hsize\centering\parindent\z@
      \vskip 0.05in\hrule height 1pt\vskip 0.22in
      {\Large\bfseries\@title\par}
      \vskip 0.2in\hrule height 1pt\vskip 0.25in
      {\@author\par}
      \vskip 0.3in}}
  \renewenvironment{abstract}{%
    \global\@topnum\z@
    \centerline{\large\bfseries Abstract}\vspace{0.5ex}
    \list{}{\leftmargin=1.2em\rightmargin=\leftmargin\listparindent\z@}\item\relax}%
    {\endlist\vspace{0.5ex}}
  \fancypagestyle{plain}{\fancyhf{}%
    \fancyfoot[C]{\thepage}}
\else
  \titleformat{\section}{\large\bfseries}{\thesection}{1em}{}
  \titleformat{\subsection}{\normalsize\bfseries}{\thesubsection}{1em}{}
  \titlespacing*{\section}{0pt}{1.6ex plus .6ex minus .2ex}{0.6ex plus .2ex}
  \titlespacing*{\subsection}{0pt}{1.2ex plus .4ex minus .2ex}{0.4ex plus .2ex}
  \renewcommand{\@maketitle}{%
    \vbox{\hsize\textwidth\linewidth\hsize\centering\parindent\z@
      \hrule height 4pt\vskip 0.25in\vskip -\parskip
      {\LARGE\bfseries\@title\par}
      \vskip 0.25in\vskip -\parskip
      \hrule height 1pt\vskip 0.3in
      {\@author\par}
      \vskip 0.3in}}
  \renewenvironment{abstract}{%
    \centerline{\large\bfseries Abstract}\vspace{0.3ex}
    \list{}{\leftmargin=0.5in\rightmargin=\leftmargin\listparindent\z@
      \parsep\parskip}\item\relax}%
    {\endlist\vspace{1ex}}
  \fancypagestyle{plain}{\fancyhf{}%
    \fancyfoot[L]{\small Preprint.}\fancyfoot[C]{\thepage}}
\fi\fi
\makeatother

\begin{document}

\maketitle

\begin{abstract}
Web platforms record user and item activity in relational databases whose schemas differ from platform to platform, which makes it difficult to transfer predictive models between them. Relational foundation models address this through large-scale cross-database pretraining, but transfer across schemas raises two challenges: learning structure that schemas share, such as entity histories connected through relationships and time, and supplying structure specific to the target database that pretraining may not cover. For the first, we introduce MoREGT, a mixture-of-relational-experts graph transformer with row and cell tokens, attention that reads row log-ages and foreign-key names, and experts routed by column metadata rather than values. For the second, a label-blind LLM agent adds validated single-row derived columns from the target schema and column statistics, without seeing labels or model scores. On 21 RelBench tasks, the MoREGT backbone improves on the Relational Transformer under the same pretraining by 2.25 points of mean AUROC and 5.83 of mean nMAE. The agent's columns raise mean AUROC from 78.45 to 79.63 and lower mean nMAE from 27.91 to 26.98, the best means on both metrics among the compared methods, including recent foundation models that pretrain on much larger corpora. These results demonstrate that transfer-oriented representations and target-specific adaptation offer a route to stronger relational foundation models.
\end{abstract}

\section{Introduction}\label{sec:intro}

\begin{figure}[htb]
    \centering
    \includegraphics[width=\figw{0.74\textwidth}{\columnwidth}]{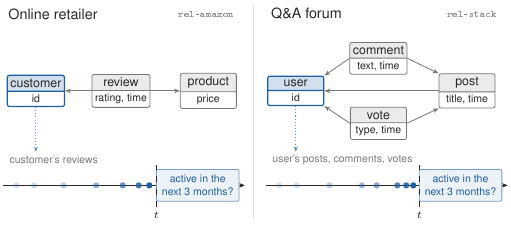}
    \caption{Two RelBench databases with no shared columns ask whether a user will stay active over the next three months. Left: online retailer \texttt{rel-amazon} (customers, products, reviews). Right: question-answering forum \texttt{rel-stack} (users, posts, comments, votes and other tables). In both, the evidence is the user's history of timestamped rows linked by foreign keys.}
    \label{fig:example}
\end{figure}

Web platforms such as online retailers, classified-ad marketplaces, question-answering forums and event services record user activity in relational databases~\citep{robinson2024relbench}. Many routine decisions on these platforms depend on predictions about users and items. An online retailer asks whether a customer will write no reviews over the next three months; a question-answering forum asks whether a user will post, comment or vote over the same period~\citep{robinson2024relbench}. The two databases share no columns: the retailer stores customers, products and reviews, while the forum stores users, posts, comments and votes. Yet both predictions rest on the same kind of evidence, a user's history of timestamped rows linked through foreign keys to other rows (Figure~\ref{fig:example}). This poses a central challenge for Web graph representation learning: transferring predictive patterns over heterogeneous temporal structured data across platforms without requiring shared entities, relationships, or database schemas.

Relational deep learning (RDL) learns such predictions directly from the database, treating rows as nodes of a temporal graph and foreign keys as edges, without hand-built features~\citep{fey2024rdl}. Recent RDL models, including graph neural networks~\citep{chen2025relgnn} and relational graph transformers~\citep{gelgt, relgt}, are usually trained for one schema and one task, so each new platform needs its own model. Relational foundation models such as Relational Transformer (RT), KumoRFM, and RT-J aim to reduce this cost by pretraining on many databases and predicting on unseen ones~\citep{ranjan2026rt,kumorfm,ranjan2026rtj}. For Web platforms with diverse relational schemas, such transfer raises two challenges. The first is learning structure that schemas share: column names and semantics differ, but entities still have histories connected through relationships and time, and a review written last week usually says more about whether a customer will stay than one written three years ago (Figure~\ref{fig:example}). A model that transfers this structure must represent elapsed time between events, treat rows as units that linked rows can attend to, and accommodate heterogeneous column semantics without forcing every column through the same feed-forward parameters. The RT backbone~\citep{ranjan2026rt} handles each indirectly. PluRel scales it with synthetic databases~\citep{kothapalli2026plurel} and RT-J with real databases, a larger model and longer contexts~\citep{ranjan2026rtj}, but both leave these representations unchanged. The second challenge is structure specific to the target database. Quantities such as code groups or rates between columns may not appear during pretraining and can be hard to learn from task labels alone~\citep{grinsztajn2022tree,jayakumar2020multiplicative}, and adding them by hand would bring back the manual feature engineering that RDL was designed to remove~\citep{fey2024rdl,robinson2024relbench}.

To this end, we introduce MoREGT, a mixture-of-relational-experts graph transformer, with one design choice for each requirement above. MoREGT represents both rows and cells as tokens, so a row token can collect information from its own cells and exchange it with the rows it links to. Attention between two rows incorporates the difference between their log-ages and the name of the foreign-key relationship that connects them. To process columns whose meaning varies across databases, MoREGT replaces shared feed-forward layers with mixtures of experts, motivated by evidence that, in a stylized setting, a mixture of experts learns a problem composed of distinct subproblems on which a dense network fails~\citep{kawata2025moe}. To enable semantically related columns across databases to share experts, they are selected using a value-free description of each cell, consisting of its column-name embedding, data type and recency. Routing therefore depends on semantics of the data (column, type and age), not on its value or hidden state. No parameter depends on a schema: per-database quantities such as column statistics and name embeddings are frozen buffers, so a single checkpoint transfers to unseen databases.

To supply target-specific structure, MoREGT is paired with an agentic loop: a label-blind LLM agent reads the target schema, the task description and column statistics and proposes validated derived columns for a given task, each a single-row formula such as a rate, a numerical band or a group of codes. The agent never accesses labels, task tables or model scores, and because each derived column reads a single row, it is exposed to the model only when its inputs are temporally visible. Unlike the task-specific aggregations over linked tables that RDL was designed to replace~\citep{fey2024rdl}, these columns leave all aggregation across rows, tables and time to the model.

Under the same leave-one-database-out pretraining as RT, the MoREGT backbone alone improves on RT by 2.25 in mean AUROC and 5.83 in mean nMAE. With the agent-derived columns added, MoREGT achieves the best mean scores for relational foundation models on the RelBench leaderboard, 79.63 AUROC and 26.98 nMAE, compared with 79.02 and 27.11 reported for RT-J~\citep{ranjan2026rtj} which is trained on a dataset with 400x more databases compared to what MoREGT is trained on. MoREGT also matches KumoRFM-2~\citep{kumorfm} on classification and improves on it and on KumoRelational on regression. Our contributions are:

\begin{itemize}[leftmargin=*,left=0pt..1em,topsep=0pt,itemsep=0pt]
    \item \textbf{Transferable relational representation learning.}
    We introduce MoREGT, a graph transformer with explicit row and cell tokens, time- and relationship-aware attention, value-free expert routing, and a relational pretraining objective designed to learn structure that transfers across schemas.

    \item \textbf{Label-blind database adaptation.}
    We develop an LLM agent that generates validated, single-row derived columns from the target database's schema and statistics, without access to task labels or model feedback.

    \item \textbf{Data-efficient cross-schema generalization.}
    With derived columns, MoREGT achieves the best mean classification and regression performance among the compared methods on RelBench with a smaller pretraining dataset, suggesting that transferable representations and target-specific adaptation can partly substitute for larger pretraining corpora.
\end{itemize}

\section{Related Work}\label{sec:related}

\noindent\textbf{Relational deep learning and foundation models.}
Relational deep learning treats a database as a temporal graph with one node per row and one edge per foreign-key link~\citep{fey2024rdl}, and RelBench provides its standard tasks~\citep{robinson2024relbench}. Models such as RelGNN~\citep{chen2025relgnn} and RelGT~\citep{relgt} build on heterogeneous graph networks~\citep{hu2020hgt} and are trained for one database, with parameters tied to its node and edge types; RelGT also gives each node a time token. Foundation models transfer across schemas instead. For single tables, CARTE embeds each cell together with its column name using a language model~\citep{kim2024carte}. For relational databases, Griffin pretrains a graph-centric model~\citep{wang2025griffin}, RT tokenizes every cell with its column name and attends along columns, rows and key links~\citep{ranjan2026rt}, KumoRFM predicts by in-context learning~\citep{kumorfm}, and PluRel and RT-J scale RT's pretraining data~\citep{kothapalli2026plurel,ranjan2026rtj}. MoREGT keeps RT's schema-free cell tokens and adds row tokens, attention that depends on log-age differences and foreign-key names, experts routed on column meaning, and an objective that masks additional cells of the seed row so that its properties must be recovered from relational context. RT-J also reports experiments with multi-cell masking~\citep{ranjan2026rtj}.

\noindent\textbf{Mixture of experts.}
Sparse mixtures of experts add parameters at a fixed cost per token~\citep{shazeer2017moe,fedus2022switch}, often alongside shared experts that process every token~\citep{dai2024deepseekmoe}, and theory shows that experts can split a task into subproblems that a dense network learns poorly~\citep{kawata2025moe}. Most routers read the token's hidden state, while hash layers route by token identity~\citep{roller2021hash} and DEMix by a document's known domain~\citep{gururangan2022demix}. Our router is likewise fixed by what the input is, a value-free signature of each cell, but unlike a token identity or a domain label, this signature is defined for columns never seen in pretraining. Because column types are long-tailed, we also replace load balancing with an orthogonality penalty on the router.

\noindent\textbf{Feature engineering and data-science agents.}
Deep feature synthesis enumerates aggregations and transformations along foreign keys~\citep{kanter2015dfs}. LLM-based methods add the meaning of columns but select features with labels: CAAFE keeps a generated feature only if it improves validation performance~\citep{hollmann2023caafe}, and FeatLLM derives rules from labelled examples~\citep{han2024featllm}. RelAgent~\citep{huang2026relagent} brings this to relational databases: an LLM agent writes multi-table SQL feature queries with temporal filters, revises them using validation scores, and trains a tree ensemble that then predicts without the LLM. Our agent sees no labels, task table or model scores, and writes only single-row formulas, which keeps derived columns free of future information by construction rather than through query filters. Its columns are added to the database for a pretrained relational model rather than replacing that model.

\section{MoREGT}
\label{sec:method}

\begin{figure*}[t]
  \centering
  \includegraphics[width=\figw{\textwidth}{0.84\textwidth}]{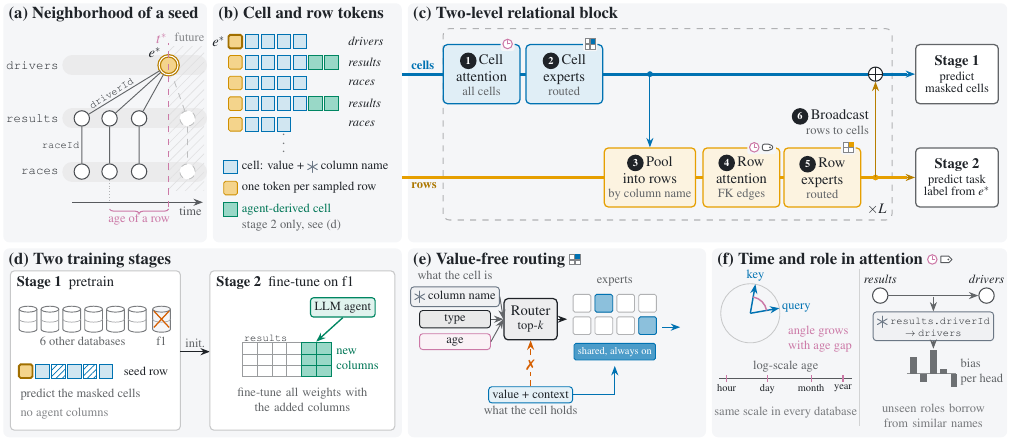}
  \caption{Overview of MoREGT. (a) Temporal neighborhood of a seed $e^*$ at time $t^*$. (b) Cell and row tokens. (c) Two-level relational block, repeated $L$ times. (d) Stage~1 pretrains on the other databases; stage~2 fine-tunes on the target database with agent-derived columns (green). (e) Expert routing reads what a cell is, never its value. (f) Attention depends on log-age and on the name of each FK role.}
  \label{fig:overview}
\end{figure*}

MoREGT predicts a property of an entity in a relational database from its temporal relational neighborhood (Figure~\ref{fig:overview}). After formalizing the setting, we describe the two parts that correspond to the two challenges of Section~\ref{sec:intro}: a schema-transferable row--cell graph transformer (Sections~\ref{sec:backbone}--\ref{sec:pretraining}) and a label-blind agent that adds target-specific derived columns (Section~\ref{sec:agent}).

\noindent\textbf{Problem setup.}
A relational database $\mathcal{D}$ is a set of tables linked by primary--foreign key (PK--FK) relationships~\citep{fey2024rdl,robinson2024relbench}. We index rows by $i$ and columns by $j$. Row $i$ belongs to table $T(i)$ and carries a timestamp $t_i$ if that table is timed, each feature column $j$ has a semantic type $\mathrm{st}(j)$ (numerical, categorical, text or timestamp), and for a feature column $j$ of $T(i)$, cell $(i,j)$ holds the value $x_{i,j}$. Following RDL, we view $\mathcal{D}$ as a heterogeneous temporal graph with one node per row. Every FK column $\texttt{col}$ of a child table $C$ that references a parent table $P$ defines a \emph{role} $\rho=(C,\texttt{col},P)$ and a pair of typed edges in both directions, so two FKs into the same parent, e.g., \texttt{buyer\_id} and \texttt{seller\_id}, remain distinct. Key columns enter the model only as graph structure, never as cell values.
A task specifies seeds $(e^*,t^*)$, an entity row and a prediction time, and asks for a label $y$ describing $e^*$ at $t^*$. For each seed, a temporal sampler expands $N_{\mathrm{hop}}$ hops from $e^*$ and keeps, per edge type and frontier node, the $N_{\mathrm{nbr}}$ most recent neighbors with $t_i\le t^*$; rows of untimed tables are not time-filtered. We write $\Delta_i=\max(0,t^*-t_i)$ for the age of row $i$ at $t^*$, in seconds, and $\tau_i=\log(1+\Delta_i)$ for its log-age. The time features $\boldsymbol{\varphi}_i\in\mathbb{R}^{2K+1}$ of row $i$ are $\sin(\omega_f\tau_i)$ and $\cos(\omega_f\tau_i)$ for $K$ fixed frequencies $\omega_f$ spaced logarithmically in $[\omega_{\min},\omega_{\max}]$, plus a flag that marks a clamped time (Appendix~\ref{app:impl}); the sine and cosine terms are set to zero for untimed rows so that an unknown age differs from $\Delta_i=0$.

\subsection{Row--Cell Relational Representation}
\label{sec:backbone}
\noindent\textbf{Cell and row embeddings.} Following RT~\citep{ranjan2026rt}, every cell of every sampled row becomes one token (Figure~\ref{fig:overview}b). The seed row comes first, cells within a row are ordered by column name, and sequences are capped at $N_{\mathrm{cell}}$ cells. An encoder shared by all columns of a semantic type maps the value $x_{i,j}$ to a vector $\mathbf{v}_{i,j}$; in particular, a categorical value $c$ is encoded through the frozen text embedding $\mathbf{n}_{j,c}$ of ``table $T$, column $j$, value $c$'', where $T$ is the table of $j$ (Appendix~\ref{app:impl} gives all encoders). The initial embedding of cell $(i,j)$, which the first block reads, adds a value component and a schema component,
\begin{equation}\label{eq:token}
\mathbf{H}^{(0)}_{i,j}=\mathbf{W}_{\mathrm{val}}\,\mathbf{v}_{i,j}+\mathbf{b}_{\mathrm{val}}+\mathbf{W}_{\mathrm{name}}\,\mathbf{n}_j+\mathbf{b}_{\mathrm{name}},
\end{equation}
where $\mathbf{n}_j\in\mathbb{R}^{d_n}$ is the frozen embedding of ``table $T$, column $j$'', and $\mathbf{W}_{\mathrm{val}}$, $\mathbf{b}_{\mathrm{val}}$, $\mathbf{W}_{\mathrm{name}}$ and $\mathbf{b}_{\mathrm{name}}$ are learned. MoREGT also keeps one embedding per sampled row, so that a row, and not only its cells, can take part in attention and be read by the prediction head. We write $\mathbf{U}^{(0)}_i$ for the initial embedding of row $i$ (Appendix~\ref{app:impl}).

\noindent\textbf{Overview.} The backbone maps the initial embeddings of one example to a prediction for its seed row. It keeps one embedding per cell, which carries content, and one per sampled row, which carries relational structure. We write $\mathbf{H}^{(\ell)}$ for the cell embeddings $\mathbf{H}^{(\ell)}_{i,j}$ and $\mathbf{U}^{(\ell)}$ for the row embeddings $\mathbf{U}^{(\ell)}_i$ after $\ell$ blocks, all in $\mathbb{R}^d$. Besides the embeddings, every block reads a context $\mathcal{G}$ that is fixed for the example and contains no cell value: the log-ages $\boldsymbol{\tau}=(\tau_i)_i$, the set $\mathcal{E}$ of sampled FK edges with the role $\rho(i,i')$ of each edge, the column-name embeddings $\mathbf{n}_j$, and the value-free signatures $\mathbf{Z}$ and $\mathbf{S}$ defined in Section~\ref{sec:moe}. The backbone applies $L$ blocks and predicts from the final embedding of the seed row:
\begin{equation}\label{eq:stack}
\begin{aligned}
(\mathbf{H}^{(\ell)},\mathbf{U}^{(\ell)})&=\mathrm{MoREGTBlock}^{(\ell)}\big(\mathbf{H}^{(\ell-1)},\mathbf{U}^{(\ell-1)};\mathcal{G}\big),\\
\hat y&=\mathrm{MLP}\big(\operatorname{LN}(\mathbf{U}^{(L)}_{e^*})\big),
\end{aligned}
\end{equation}
where $\ell=1,\dots,L$, each block has its own parameters, and $\operatorname{LN}$ is LayerNorm.

\noindent\textbf{Inside a block.} $\mathrm{MoREGTBlock}^{(\ell)}$ takes the cell embeddings $\mathbf{H}^{(\ell-1)}$, the row embeddings $\mathbf{U}^{(\ell-1)}$ and the context $\mathcal{G}$, and returns new embeddings $\mathbf{H}^{(\ell)}$ and $\mathbf{U}^{(\ell)}$ of the same sizes. It applies the six sublayers numbered in Figure~\ref{fig:overview}c: two update the cells, one pools the cells into their rows, two update the rows, and one broadcasts the rows back to their cells:
\begin{align}
\mathbf{H}'&=\mathbf{H}^{(\ell-1)}+\mathrm{CellAttn}\big(\operatorname{RMSNorm}(\mathbf{H}^{(\ell-1)});\boldsymbol{\tau}\big),\notag\\
\mathbf{H}''&=\mathbf{H}'+\mathrm{CellMoE}\big(\operatorname{RMSNorm}(\mathbf{H}');\mathbf{Z}\big),\notag\\
\mathbf{U}'&=\mathbf{U}^{(\ell-1)}+\mathrm{Pool}\big(\mathbf{U}^{(\ell-1)},\mathbf{H}'';\mathbf{S},\{\mathbf{n}_j\}\big),\notag\\
\mathbf{U}''&=\mathbf{U}'+\mathrm{RowAttn}\big(\operatorname{RMSNorm}(\mathbf{U}');\boldsymbol{\tau},\mathcal{E}\big),\notag\\
\mathbf{U}^{(\ell)}&=\mathbf{U}''+\mathrm{RowMoE}\big(\operatorname{RMSNorm}(\mathbf{U}'');\mathbf{S}\big),\notag\\
\mathbf{H}^{(\ell)}_{i,j}&=\mathbf{H}''_{i,j}+\mathbf{W}_b\,\mathbf{U}^{(\ell)}_i,\label{eq:block}
\end{align}
where the arguments after the semicolons are the parts of $\mathcal{G}$ that each sublayer reads, $\operatorname{RMSNorm}$ is applied to each embedding separately, and $\mathbf{W}_b$ is learned. Every update is residual. The attention and expert sublayers are pre-norm, while the two transfers between levels, pooling and broadcast, read the embeddings as they are. Three components are specific to MoREGT: attention scores that depend on the log-ages of the rows, a row-attention bias computed from the name of the FK role that links two rows (both in Section~\ref{sec:attention} and Figure~\ref{fig:overview}f), and experts routed on value-free signatures (Section~\ref{sec:moe} and Figure~\ref{fig:overview}e). We group the sublayers by their function below, while Figure~\ref{fig:overview}c gives their execution order; Appendix~\ref{app:impl} gives the remaining specifications.

\noindent\textbf{\stepnum{3} Pool into rows.} Row attention moves information along FK edges, and FK edges link rows, not cells, so before that step each row needs a summary of its own cells as they stand after the cell sublayers. $\mathrm{Pool}$ computes this summary by multi-head cross-attention from each row to its own cells. The queries come from the row's embedding and signature $[\mathbf{U}^{(\ell-1)}_i;\mathbf{S}_i]$, the keys from the frozen column-name embeddings $\mathbf{n}_j$, and the values from the cell embeddings $\mathbf{H}''_{i,j}$. Because the keys are column names rather than cell embeddings, each head selects cells by what their columns mean, regardless of their values. Because the names come from a frozen encoder, the columns of an unseen database receive keys in the same space as the pretraining columns.

\noindent\textbf{\stepnum{6} Broadcast to cells.} The last line of Eq.~\eqref{eq:block} returns what each row gathered along FK edges to its cells, adding a projection of the row embedding to every cell of the row. The next block's cell attention, and after block $L$ the pretraining heads, read these cell embeddings.

Unlike RT, which represents rows only through their cells, MoREGT retains row tokens for FK-level message passing and downstream prediction.

\subsection{Time- and Relationship-Aware Attention}
\label{sec:attention}
\noindent\textbf{\stepnum{1} Cell attention.} Whether a related row informs a prediction at $t^*$ can depend on how long before $t^*$ it occurred relative to other rows, and a pretrained model must apply to databases that cover very different time spans. RT embeds a timestamp as a cell value standardized over the whole database~\citep{ranjan2026rt}, so the time between two rows must be inferred by comparing two values whose scale is set by that span. MoREGT instead puts the log-ages of the two rows into every attention score. $\mathrm{CellAttn}$ is multi-head attention among all cells of the example with no structural mask, so content mixes freely across rows and tables. Following rotary embeddings~\citep{su2024roformer}, $K$ pairs of dimensions of the query and key of every token of row $i$ are rotated by the angles $\omega_f\tau_i$, $f=1,\dots,K$, with the frequencies of $\boldsymbol{\varphi}_i$. Rotating a query by one angle and a key by another leaves only the difference of the angles in their dot product, so in one head the score of cell $p=(i,j)$ for cell $p'=(i',j')$, which $\mathrm{CellAttn}$ uses between cells and $\mathrm{RowAttn}$ uses between rows, is
\begin{equation}\label{eq:rope}
\sigma_{pp'}=\frac{1}{\sqrt{d_h}}\Big(\bar{\mathbf{q}}_p^{\top}\bar{\mathbf{k}}_{p'}+\sum_{f=1}^{K}\mathbf{q}_p[f]^{\top}\mathbf{R}\big(\omega_f(\tau_{i'}-\tau_i)\big)\,\mathbf{k}_{p'}[f]\Big),
\end{equation}
where $d_h$ is the head width, $\bar{\mathbf{q}}_p$ and $\bar{\mathbf{k}}_{p'}$ are the unrotated parts of the query of $p$ and the key of $p'$, $\mathbf{q}_p[f],\mathbf{k}_{p'}[f]\in\mathbb{R}^2$ are their $f$-th rotated pairs, and $\mathbf{R}(\theta)$ is the planar rotation by the angle $\theta$. Two learned offsets mark pairs in which a row is untimed or its time was clamped (Appendix~\ref{app:impl}). Otherwise time enters the score only through $\tau_{i'}-\tau_i=\log\frac{1+\Delta_{i'}}{1+\Delta_i}$. Because ages are measured in seconds, the $+1$ is negligible and this difference is the log-ratio of the two ages: one day versus two days is as far apart as one year versus two years (both $\approx\log 2$), whatever time span the database covers. Cells carry no row identifier, so in the first block cells of rows with equal ages are not told apart by row; from the second block on, broadcast \stepnum{6} adds each row's embedding to its cells.

\noindent\textbf{\stepnum{4} Row attention.} Rows linked through different FKs play different parts: the buyer and the seller of an item are both users but relate to the item differently. The RDL baseline~\citep{fey2024rdl} passes messages with separate parameters per edge type, and such parameters do not exist for an edge type unseen in pretraining. MoREGT instead computes an attention bias from the name of the role that links two rows (Figure~\ref{fig:overview}f). $\mathrm{RowAttn}$ is multi-head attention in which row $i$ attends to itself and to the set $\mathcal{N}(i)=\{i':(i,i')\in\mathcal{E}\}$ of rows linked to it by a sampled FK edge. In each head $m$, the score of row $i$ for row $i'$ is the time-aware score of Eq.~\eqref{eq:rope}, computed from row queries and keys, plus a bias that, for $i'\in\mathcal{N}(i)\cup\{i\}$, is
\begin{equation}\label{eq:rolebias}
\beta^m_{ii'}=\mathbf{a}_m^{\top}\mathbf{W}_{\mathrm{att}}\mathbf{n}^{\mathrm{role}}_{\rho(i,i')}+b_{m,\mathrm{dir}(i,i')},
\end{equation}
where $\mathbf{n}^{\mathrm{role}}_{\rho(i,i')}$ is the frozen embedding of ``table $C$ column \texttt{col} references table $P$'' for the role $\rho(i,i')=(C,\texttt{col},P)$ of the edge, $\mathbf{W}_{\mathrm{att}}$ is a projection shared by the heads, $\mathbf{a}_m$ and $b_{m,\cdot}$ are learned per head, and $\mathrm{dir}(i,i')$ is \emph{parent} if row $i$ holds the FK that references $i'$, \emph{child} if $i'$ holds the FK that references $i$, and \emph{self} if $i'=i$, in which case the role term is zero. Because each FK column defines its own role, the buyer and the seller of an item receive different biases. Because the bias is computed from the role's name rather than learned per role, a role unseen in pretraining receives a bias close to that of similarly named roles.

Unlike RT, which compares standardized timestamp values~\citep{ranjan2026rt}, and RDL models, which learn separate parameters per edge type~\citep{fey2024rdl}, MoREGT puts relative log-age into both cell and row attention and derives edge-specific biases from FK-role names without schema-specific parameters.

\subsection{Value-Free Mixture-of-Relational-Experts}
\label{sec:moe}
\noindent\textbf{Value-free signatures.} A signature carries the semantics of what a cell or row is. Signatures are the only input of the expert routers (Eq.~\eqref{eq:cell-moe}), so they decide which parameters process each cell and row. Because they hold no value, this decision is fixed by the schema and by age, and because they are built from frozen name embeddings, a new database receives signatures in the same space as the pretraining databases. Both kinds encode age through the time features $\boldsymbol{\varphi}_i$ of row $i$ defined in the problem setup. The signature of cell $(i,j)$ combines the meaning of its column, its semantic type and the age of its row, so cells of one column can take different routes in recent and old rows:
\begin{equation}\label{eq:cell-sig}
\mathbf{Z}_{i,j}=\operatorname{RMSNorm}\big(\mathbf{W}_z\mathbf{n}_j+\mathbf{E}_{\mathrm{st}}[\mathrm{st}(j)]+\mathbf{W}_\tau\boldsymbol{\varphi}_i\big),
\end{equation}
where $\mathbf{Z}_{i,j}\in\mathbb{R}^{d_s}$, $\mathbf{W}_z$ and $\mathbf{W}_\tau$ are learned projections, and $\mathbf{E}_{\mathrm{st}}$ holds one learned embedding per semantic type. The signature of row $i$ says what kind of row it is and how it relates to the seed, through its table, the FK role through which it is first reached from the seed, its hop distance and its age:
\begin{equation}\label{eq:row-sig}
\begin{aligned}
\mathbf{S}_i=\operatorname{RMSNorm}\big(&\mathbf{W}_T\mathbf{n}^{\mathrm{tab}}_{T(i)}+\mathbf{W}_\rho\mathbf{n}^{\mathrm{role}}_{\rho_0(i)}\\
&+\mathbf{E}_{\mathrm{hop}}[\mathrm{hop}(i)]+\mathbf{W}'_{\tau}\boldsymbol{\varphi}_i\big),
\end{aligned}
\end{equation}
where $\mathbf{n}^{\mathrm{tab}}_{T}$ is the frozen embedding of the name of table $T$, $\mathbf{n}^{\mathrm{role}}_{\rho}$ is the role embedding of Eq.~\eqref{eq:rolebias}, $\rho_0(i)$ is the role through which row $i$ is first reached (the role term is zero for the seed row), $\mathbf{E}_{\mathrm{hop}}$ holds one learned embedding per hop, and $\mathbf{W}_T$, $\mathbf{W}_\rho$ and $\mathbf{W}'_\tau$ are learned projections.

\noindent\textbf{\stepnum{2} Cell experts.} $\mathrm{CellMoE}$ is a mixture of experts whose router reads only the signature $\mathbf{Z}_{i,j}$, while the experts transform the normalized cell embedding $\hat{\mathbf{H}}_{i,j}=\operatorname{RMSNorm}(\mathbf{H}'_{i,j})$ (Figure~\ref{fig:overview}e). Its output is added to the cell embedding in the second line of Eq.~\eqref{eq:block}:
\begin{equation}\label{eq:cell-moe}
\begin{aligned}
\mathrm{CellMoE}(\hat{\mathbf{H}}_{i,j};\mathbf{Z}_{i,j})&=\sum_{e\in\mathcal{T}_{i,j}}g_{i,j,e}\,\mathrm{E}_e(\hat{\mathbf{H}}_{i,j})+\sum_{e=1}^{N_{\mathrm{s}}}\mathrm{E}^{\mathrm{sh}}_e(\hat{\mathbf{H}}_{i,j}),\\
\mathbf{g}_{i,j}&=\operatorname{softmax}\big(\operatorname{TopK}_{k_{\mathrm{r}}}(\mathbf{W}_g\mathbf{Z}_{i,j})\big),
\end{aligned}
\end{equation}
where $\mathbf{W}_g$ is the router, which scores the $N_{\mathrm{r}}$ routed experts $\mathrm{E}_e$; $\operatorname{TopK}_{k_{\mathrm{r}}}$ keeps the $k_{\mathrm{r}}$ largest scores and sets the others to $-\infty$; $\mathcal{T}_{i,j}$ is the set of experts it keeps; and $\mathrm{E}^{\mathrm{sh}}_1,\dots,\mathrm{E}^{\mathrm{sh}}_{N_{\mathrm{s}}}$ are always-active shared experts~\citep{dai2024deepseekmoe}. All experts are feed-forward networks. The cell router is regularized only by an orthogonality penalty on its expert directions, with no load-balancing loss (Appendix~\ref{app:impl}).

We route on signatures because a dense feed-forward layer shared by every column of every database, as in RT~\citep{ranjan2026rt}, must fit one pooled problem, which each database added to pretraining makes more heterogeneous. For data made of distinct subproblems, a dense network trained by gradient descent can fail to find this structure, while a mixture of experts splits the task and learns each subproblem~\citep{kawata2025moe}. In MoREGT the subproblems are observed rather than latent: they are given by the column, type and age of each cell. The router therefore reads what a cell is, and the experts transform what it holds after attention has mixed in context. A routed expert's parameters enter the sublayer only at the cells routed to it, so it is trained only on those cells, and this assignment depends neither on values nor on which neighbors were sampled. For the same reason, the route cannot encode the label.

\noindent\textbf{\stepnum{5} Row experts.} $\mathrm{RowMoE}$ applies Eq.~\eqref{eq:cell-moe} to the normalized row embeddings, with its own experts and a cosine router~\citep{chi2022representation} on the row signature $\mathbf{S}_i$, so that rows of different tables, roles and hops can use different experts. Its router is regularized by the same orthogonality penalty and, unlike the cell router, by a load-balancing term (Appendix~\ref{app:impl}).

Unlike conventional content-dependent MoE routing~\citep{shazeer2017moe,fedus2022switch}, expert assignment uses schema metadata and age, allowing related columns in an unseen database to select the same pretrained experts.

\subsection{Relational Pretraining and Transfer}
\label{sec:pretraining}
\noindent\textbf{Relational pretraining objective.} Every table with a maskable column (numerical with nonzero variance, or categorical with at least two classes) supplies seeds; a timed row uses its own timestamp as $t^*$, and a static row draws $t^*$ from the training window. RT masks a single target cell and predicts it from its context~\citep{ranjan2026rt}, which leaves the other cells of that row visible, so a column that is close to a function of its siblings can be recovered without the relational context. We mask one maskable cell of the seed row, as in RT, and each other maskable cell of the seed row with probability $p_{\mathrm{mask}}$, and we never mask cells of context rows. The objective thus has the form of a downstream task, predicting a property of an entity at $t^*$ from its relational past. Numerical targets are regressed with a Huber loss from $\tilde{\mathbf{H}}_{i,j}=\operatorname{RMSNorm}(\mathbf{H}^{(L)}_{i,j})$, and categorical targets are scored against the column's own frozen category embeddings, $o_{i,j,c}=(\mathbf{W}_{\mathrm{cat}}\tilde{\mathbf{H}}_{i,j}+\mathbf{b}_{\mathrm{cat}})^{\top}\mathbf{n}_{j,c}$, so the output space has no schema-specific parameters either. The pretraining loss is $\mathcal{L}_{\mathrm{num}}+\lambda_{\mathrm{cat}}\mathcal{L}_{\mathrm{cat}}+\lambda_{\mathrm{aux}}\mathcal{L}_{\mathrm{aux}}$, where $\mathcal{L}_{\mathrm{aux}}$ collects the routing regularizers of both levels over all blocks (Appendix~\ref{app:impl}).

Unlike single-cell reconstruction in RT~\citep{ranjan2026rt}, additional seed-row masking reduces the opportunity to reconstruct a target from its sibling cells alone. RT-J also reports experiments with multi-cell masking~\citep{ranjan2026rtj}; in MoREGT the extra masking is confined to the seed row, and context rows are never masked.

\noindent\textbf{Transfer to unseen databases.} No parameter of MoREGT has a shape that depends on a database: every per-database quantity (column statistics and name, category and role embeddings) is a frozen buffer rebuilt for each database, so one checkpoint applies to any schema. For each RelBench database $\mathcal{D}$, we pretrain a trunk $\theta^{(-\mathcal{D})}$ on all other databases of the benchmark, so it has never seen a row, a column name or the schema of $\mathcal{D}$; all trunks share one recipe and use no rows past the validation timestamp. Fine-tuning on a task of $\mathcal{D}$ starts from $\theta^{(-\mathcal{D})}$, replaces the reconstruction heads by the prediction head of Eq.~\eqref{eq:stack}, and trains all parameters with binary cross-entropy or, for regression, an $\ell_1$ loss on standardized targets, whose minimizer, the conditional median, is MAE-optimal; in both cases the loss includes $\lambda_{\mathrm{aux}}\mathcal{L}_{\mathrm{aux}}$. Appendix~\ref{app:impl} gives all hyperparameters and the optimization and model-selection procedures.

As in RT~\citep{ranjan2026rt}, and unlike RDL models whose parameters are tied to one database's node and edge types~\citep{chen2025relgnn,relgt}, this two-stage procedure retains a single schema-independent parameterization: pretraining learns transferable relational structure, while fine-tuning adapts that structure to an unseen database.

\subsection{Label-Blind Agentic Loop for Derived Columns}
\label{sec:agent}

\begin{figure}[t]
  \centering
  \includegraphics[width=\figw{0.56\textwidth}{\columnwidth}]{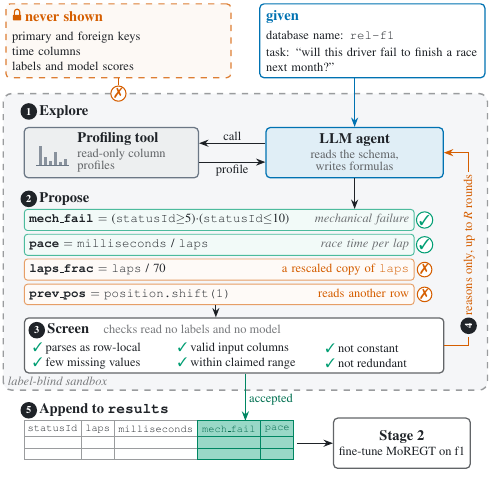}
  \caption{Agentic loop for derived columns on \texttt{rel-f1}. A label-blind LLM agent profiles the data, proposes row-local formulas and revises them from screening feedback. Accepted formulas become new columns, used only in stage~2 (Figure~\ref{fig:overview}d). The proposals shown are illustrative.}
  \label{fig:agent}
\end{figure}

\noindent\textbf{Why derived columns, and why an agent.} Some quantities that matter for a task are conventions of one database: in Formula~1, status codes 5 to 10 denote mechanical failures, and race time divided by completed laps gives a driver's pace.
Each takes one line to write, but code groups and bands are irregular functions that neural networks learn poorly~\citep{grinsztajn2022tree}, and rates are interactions between columns~\citep{jayakumar2020multiplicative}. A derived column hands such a quantity to the model as one extra cell per row.
Derived columns are not free. Each is a function of its own row, so it adds no information about the target beyond that row, and each costs one token per sampled row under the cap $N_{\mathrm{cell}}$ (Appendix~\ref{app:impl}). Useful columns are therefore few and depend on what the columns mean, which favors an LLM agent that reads the schema over enumerating transformations~\citep{kanter2015dfs}.

\noindent\textbf{A label-blind agent.} We use a label-blind LLM agent (Figure~\ref{fig:agent}) that receives the database name and a one-line task description; its only action is a read-only profiling tool that returns schemas, column profiles (null rate, cardinality, quantiles, moments and frequent values) and key candidates inferred from the data. All declared primary keys, foreign keys and time columns are hidden; the agent never sees the task table, labels or model scores. This enables agent columns to be produced before any labels exist. Each tool call is parsed from the transcript, and any run with a call outside the allowlist is discarded.

\begin{definition}[Row-local expression]\label{def:grammar}
A row-local expression on table $T$ is built from the columns of $T$, numeric constants, arithmetic operators (including floor division, modulo and powers), unary signs, comparisons (which evaluate to $0$ or $1$), and the element-wise functions $\mathrm{abs}$, $\log$, $\mathrm{log1p}$, $\sqrt{\cdot}$, $\min$ and $\max$. Attribute access, subscripts, Boolean connectives, aggregation, grouping, sorting, shifting, and references to other tables or other proposals are excluded.
\end{definition}

\noindent\textbf{Screening and feedback.} The agent proposes named row-local expressions $x_{i,j}=f_j(x_{i,j_1},\dots,x_{i,j_m})$, each defining a new column $j$ and coming with a description, units, a claimed range and a rationale. A candidate is rejected if it fails to parse or evaluate, references a missing or forbidden column, is constant, infinite or entirely missing, has a missing rate more than $\eta$ above that of its inputs, places less than a fraction $q$ of its values inside its own claimed range, or has absolute correlation above $r_{\max}$ with any numerical column of its table, including earlier proposals. Rejection reasons are returned to the agent for up to $R$ rounds, and the feedback contains no model score. Accepted expressions are evaluated over their tables, missing and infinite results are set to $0$, and each result becomes an ordinary numerical column with its own statistics and name embedding, which MoREGT processes like any native column.

\noindent\textbf{Temporal visibility by construction.} The sampler admits a timed row only if $t_i\le t^*$. A derived value is a function of one row, so it is visible to the model exactly when its inputs are. Aggregation, shifting and cross-table references, which could read rows from after $t^*$, are excluded from the grammar for this reason.

Unlike label-guided feature search~\citep{hollmann2023caafe,han2024featllm,huang2026relagent} or multi-table aggregations~\citep{kanter2015dfs}, this adaptation requires no downstream labels or model feedback and preserves row-local temporal visibility.

\section{Experiments}\label{sec:exp}
In this section, we first describe the experimental setup used to train MoREGT. We then outline baselines followed by results and ablation studies.

  \begin{table*}[t]
  \centering
  \caption{Entity classification on RelBench: TEST AUROC (\%; higher is better). \textbf{Bold}: best; \underline{underline}: second best.   $^{\ddagger}$From \citet{kumorfm}. RT reproduced with checkpoints from \citet{ranjan2026rt}, other baselines from the RelBench leaderboard.}
  \label{tab:board-clf}
  \footnotesize
  \setlength{\tabcolsep}{3pt}
  \begin{adjustbox}{max width=\textwidth}
  \begin{tabular}{@{}ll ccccccccc@{}}
  \toprule
   & & \multicolumn{2}{c}{\textit{From scratch}} & \multicolumn{3}{c}{\textit{Pretrained + fine-tuned}} & \multicolumn{2}{c}{\textit{In-context}} & \multicolumn{2}{c}{\textit{Ours (MoREGT)}} \\
  \cmidrule(lr){3-4} \cmidrule(lr){5-7} \cmidrule(lr){8-9} \cmidrule(lr){10-11}
  \textbf{Dataset} & \textbf{Task} & \textbf{RelGT} & \textbf{GelGT} & \textbf{RT} & \textbf{RT-J} & \textbf{RT-PluRel} & \textbf{KumoRel.} & \textbf{KumoRFM-2$^{\ddagger}$} & \textbf{\begin{tabular}[b]{@{}c@{}}MoREGT\\(w/o agent)\end{tabular}} &
  \textbf{\begin{tabular}[b]{@{}c@{}}MoREGT\\(with agent)\end{tabular}} \\
  \midrule
  \texttt{amazon} & item-churn & 82.5 & 81.9 & \textbf{83.4} & \underline{83.2} & \underline{83.2} & 82.5 & 82.2 & 81.36\pmstd{0.38} & 81.71\pmstd{0.22} \\
   & user-churn & 70.4 & 69.5 & 70.8 & \textbf{71.5} & \underline{71.3} & 69.5 & 69.1 & 69.44\pmstd{0.31} & 70.17\pmstd{0.15} \\
  \addlinespace[2pt]
  \texttt{avito} & user-clicks & \underline{68.3} & \textbf{68.4} & 59.0 & 66.8 & 65.9 & 68.1 & 67.4 & 64.69\pmstd{2.62} & 67.79\pmstd{1.37} \\
   & user-visits & 66.8 & \underline{67.1} & 65.2 & 66.9 & 66.8 & 67.0 & \textbf{69.4} & 65.26\pmstd{1.50} & 66.65\pmstd{0.36} \\
  \addlinespace[2pt]
  \texttt{event} & user-ignore & 81.6 & \underline{88.4} & 72.4 & 85.6 & 80.9 & 85.1 & \textbf{90.8} & 85.67\pmstd{5.76} & 86.31\pmstd{3.38} \\
   & user-repeat & 76.1 & \textbf{84.1} & 69.9 & 79.8 & 80.3 & 81.6 & 81.7 & \underline{82.84}\pmstd{1.94} & 82.63\pmstd{2.42} \\
  \addlinespace[2pt]
  \texttt{f1} & driver-dnf & 75.9 & 75.6 & 84.2 & 82.8 & 82.1 & \underline{84.3} & \textbf{84.6} & 83.90\pmstd{0.25} & 84.11\pmstd{0.39} \\
   & driver-top3 & 83.5 & 85.9 & 91.9 & 91.7 & 90.0 & 90.8 & \underline{92.2} & \textbf{92.27}\pmstd{0.92} & 92.06\pmstd{0.76} \\
  \addlinespace[2pt]
  \texttt{hm} & user-churn & 69.3 & 68.2 & 70.5 & 70.5 & \underline{70.6} & 69.9 & 69.3 & 68.13\pmstd{0.35} & \textbf{70.82}\pmstd{0.06} \\
  \addlinespace[2pt]
  \texttt{stack} & user-badge & 86.3 & 87.5 & 88.7 & 89.2 & \underline{89.3} & 88.1 & 87.2 & 88.31\pmstd{0.28} & \textbf{89.40}\pmstd{0.12} \\
   & user-engagement & 90.5 & \underline{90.7} & 90.2 & 90.6 & \textbf{90.9} & 90.2 & 89.4 & 90.65\pmstd{0.37} & \textbf{90.90}\pmstd{0.14} \\
  \addlinespace[2pt]
  \texttt{trial} & study-outcome & 68.6 & 72.5 & 68.2 & 69.6 & 71.2 & \textbf{74.0} & 72.0 & 68.93\pmstd{5.56} & \underline{73.05}\pmstd{1.72} \\
  \midrule
  \multicolumn{2}{@{}l}{\textit{Mean}} & 76.65 & 78.32 & 76.20 & 79.02 & 78.54 & 79.26 & \underline{79.60} & 78.45 & \textbf{79.63} \\
  \bottomrule
  \end{tabular}
  \end{adjustbox}
  \end{table*}

  \begin{table*}[t]
  \centering
  \caption{Entity regression on RelBench: TEST nMAE $=$ MAE$/\sigma_{\text{train}}$ (\%; lower is better). \textbf{Bold}: best; \underline{underline}: second best.   $^{\ddagger}$From \citet{kumorfm}; its published MAE converted to nMAE with the same train-label $\sigma$. RT reproduced with checkpoints from \citet{ranjan2026rt}, other baselines from the RelBench leaderboard.}
  \label{tab:board-reg}
  \footnotesize
  \setlength{\tabcolsep}{3pt}
  \begin{adjustbox}{max width=\textwidth}
  \begin{tabular}{@{}ll ccccccccc@{}}
  \toprule
   & & \multicolumn{2}{c}{\textit{From scratch}} & \multicolumn{3}{c}{\textit{Pretrained + fine-tuned}} & \multicolumn{2}{c}{\textit{In-context}} & \multicolumn{2}{c}{\textit{Ours (MoREGT)}} \\
  \cmidrule(lr){3-4} \cmidrule(lr){5-7} \cmidrule(lr){8-9} \cmidrule(lr){10-11}
  \textbf{Dataset} & \textbf{Task} & \textbf{RelGT} & \textbf{GelGT} & \textbf{RT} & \textbf{RT-J} & \textbf{RT-PluRel} & \textbf{KumoRel.} & \textbf{KumoRFM-2$^{\ddagger}$} & \textbf{\begin{tabular}[b]{@{}c@{}}MoREGT\\(w/o agent)\end{tabular}} &
  \textbf{\begin{tabular}[b]{@{}c@{}}MoREGT\\(with agent)\end{tabular}} \\
  \midrule
  \texttt{amazon} & item-ltv & 8.3 & 8.3 & 12.4 & \textbf{7.2} & \underline{7.3} & 7.5 & 8.0 & 7.95\pmstd{0.18} & 7.64\pmstd{0.10} \\
   & user-ltv & 24.8 & 25.2 & 27.3 & \underline{24.3} & 24.6 & 25.0 & \textbf{24.2} & 25.91\pmstd{0.36} & 25.02\pmstd{0.18} \\
  \addlinespace[2pt]
  \texttt{avito} & ad-ctr & 36.1 & \underline{35.3} & 41.9 & 36.0 & 37.9 & \textbf{33.2} & 35.5 & 35.49\pmstd{0.23} & 35.52\pmstd{0.43} \\
  \addlinespace[2pt]
  \texttt{event} & user-attendance & 32.7 & \textbf{30.6} & 56.6 & 31.6 & 31.6 & 31.7 & \underline{30.7} & 30.92\pmstd{0.26} & 31.09\pmstd{0.62} \\
  \addlinespace[2pt]
  \texttt{f1} & driver-position & 55.8 & 53.3 & \underline{39.7} & \textbf{37.7} & \underline{39.7} & 43.7 & 40.6 & \textbf{37.67}\pmstd{1.20} & \textbf{37.70}\pmstd{1.98} \\
  \addlinespace[2pt]
  \texttt{hm} & item-sales & 10.8 & 11.2 & 13.3 & 7.6 & \underline{7.5} & 8.0 & \textbf{6.9} & 8.27\pmstd{0.28} & 7.92\pmstd{0.13} \\
  \addlinespace[2pt]
  \texttt{stack} & post-votes & \underline{12.8} & 12.9 & 14.9 & \underline{12.8} & 13.6 & \textbf{12.5} & \textbf{12.5} & 14.15\pmstd{1.38} & 13.30\pmstd{0.00} \\
  \addlinespace[2pt]
  \texttt{trial} & site-success & \underline{68.6} & \textbf{68.4} & 83.6 & 76.5 & 75.2 & 85.2 & 91.0 & 77.64\pmstd{6.01} & 74.84\pmstd{5.56} \\
   & study-adverse & 13.0 & 12.1 & 14.0 & 10.3 & 10.7 & \textbf{9.4} & 12.8 & 13.21\pmstd{0.13} & \underline{9.75}\pmstd{0.27} \\
  \midrule
  \multicolumn{2}{@{}l}{\textit{Mean}} & 29.20 & 28.59 & 33.74 & \underline{27.11} & 27.57 & 28.47 & 29.13 & 27.91 & \textbf{26.98} \\
  \bottomrule
  \end{tabular}
  \end{adjustbox}
  \end{table*}
\noindent\textbf{Experimental setup.}
We evaluate on the 21 entity tasks of the seven RelBench databases~\citep{robinson2024relbench}: 12 binary classification tasks scored by AUROC and 9 regression tasks scored by nMAE (MAE divided by the standard deviation of the training labels). We use the official temporal train/validation/test splits and report test scores. Pretraining is leave-one-database-out: for each target database, a 134M-parameter MoREGT (10 blocks, width 256, mixture-of-experts feed-forward layers with 8 cell-level and 4 row-level experts) is pretrained with masked-cell prediction on the other six databases, so pretraining never sees the target database. Column names are embedded once with a frozen Qwen3-Embedding-4B and text values with MiniLM. Each fold is pretrained on 4 NVIDIA H100 GPUs for 18 hours (batch size 64, sequence length 1{,}024, learning rate  $3\times10^{-4}$), keeping the checkpoint with the best validation masked-cell score; Appendix Table~\ref{tab:hparams} lists all hyperparameters. For each task we fine-tune the full model on one H100 with past labels given as context, following RT. We fine-tune the pre-trained model for 3 seeds and early stopping on validation. For MoREGT (with agent), an LLM agent (Claude Opus~5) also proposes 8--14 derived columns per task (Appendix Table~\ref{tab:agent-columns}), stating each column's expected units and range before its values are computed. The code is available at \url{\codeurl}.

\noindent\textbf{Baselines.}
We compare against three groups of methods. \emph{Trained from scratch:} RelGT~\citep{relgt} and GelGT~\citep{gelgt}, graph transformers trained separately on each task without pretraining. \emph{Pretrained and fine-tuned:} the Relational Transformer (RT)~\citep{ranjan2026rt}, which uses the same cell-token input as our model and is pretrained on the other RelBench databases before fine-tuning on the target task, making it the most direct comparison to MoREGT (w/o agent); and RT-J and RT-PluRel, RT variants with larger pretraining corpora~\citep{ranjan2026rtj,kothapalli2026plurel}. \emph{In-context:} KumoRelational and KumoRFM-2~\citep{kumorfm}, relational foundation models that predict from labelled examples in context, without task-specific training. RT results are copied from \citet{ranjan2026rt} where reported and otherwise reproduced from its released checkpoints. KumoRFM-2 results are taken from \citet{kumorfm}, with its published MAE converted to nMAE; all other baselines come from the RelBench leaderboard.

\subsection{Results}\label{sec:res}
Tables~\ref{tab:board-clf} and~\ref{tab:board-reg} compare MoREGT with graph transformers trained from scratch (RelGT, GelGT), pretrained and fine-tuned relational transformers (RT, RT-J, RT-PluRel), and in-context relational foundation models (KumoRelational, KumoRFM-2). MoREGT (with agent) attains the best mean on both: 79.63 AUROC on the 12 classification tasks, against 79.60 for KumoRFM-2 and 79.02 for RT-J, and 26.98\% nMAE on the 9 regression tasks, against 27.11\% for RT-J and 29.13\% for KumoRFM-2. Its pretraining covers only the six RelBench databases other than the target, compared with roughly 2,400 for RT-J. The lead comes from consistency rather than per-task wins. MoREGT (with agent) is best or tied for best on 4 of the 21 tasks (\texttt{hm} user-churn, both \texttt{stack} classification tasks, \texttt{f1} driver-position) and within one point of the best on the \texttt{f1} and \texttt{trial} classification tasks. On regression, its largest deficit to the best baseline (6.4 nMAE on \texttt{trial} site-success) is smaller than that of any other method (8.1 for RT-J, 22.6 for KumoRFM-2). It trails the best baseline by 1--3 points on \texttt{amazon} and \texttt{avito}, and by 4.5 AUROC on \texttt{event} user-ignore. Without agent-derived columns, MoREGT (w/o agent) (78.45 AUROC / 27.91\% nMAE) outperforms both from-scratch baselines and RT. The classification gap to KumoRFM-2 (0.03 AUROC) is within seed variation; the two models differ chiefly in setting, since KumoRFM-2 predicts in-context without task training while MoREGT is fine-tuned per task. Appendix Figure~\ref{fig:latency} reports inference time.

\noindent\textbf{Component Ablations.}
To assess whether the architectural components benefit cross-database transfer, we ablate MoREGT under leave-one-database-out pretraining on rel-f1 (Table~\ref{tab:lodo-ablation}) by removing the introduced components. Each variant is pretrained from scratch under the same leave-one-database-out (LODO) protocol and fine-tuned with identical task data and agent-derived columns. The full model outperforms dense FFNs on all three tasks, improving AUROC by 0.64--1.66 points and reducing nMAE by 0.67 points. Replacing value-free routing with hidden-state routing, at comparable parameter count, also degrades all three tasks, supporting routing based on column semantics rather than token representations. Removing FK-role biases degrades two tasks, with a small improvement on driver-top3.
\begin{table}[t]
  \centering
  \caption{Component ablation under leave-one-database-out pretraining.
  TEST AUROC for classification and nMAE for regression (\%);
  mean $\pm$ std over 3 seeds.}
  \label{tab:lodo-ablation}
  \footnotesize
  \setlength{\tabcolsep}{3pt}
  \begin{tabular}{@{}lccccc@{}}
  \toprule
  \textbf{Task} & \textbf{Full model} & \textbf{Dense FFN} &
  \textbf{\begin{tabular}[b]{@{}c@{}}Hidden\\routing\end{tabular}} &
  \textbf{\begin{tabular}[b]{@{}c@{}}No FK-role\\bias\end{tabular}} \\
  \midrule
  driver-dnf $\uparrow$ &
  \textbf{84.11}\pmstd{0.39} &
  82.45\pmstd{1.07} &
  82.82\pmstd{0.62} &
  82.63\pmstd{1.13} \\

  driver-top3 $\uparrow$ &
  92.06\pmstd{0.76} &
  91.42\pmstd{0.15} &
  92.00\pmstd{0.26} &
  \textbf{92.33}\pmstd{0.16} \\

  driver-position $\downarrow$ &
  \textbf{37.70}\pmstd{1.98} &
  38.37\pmstd{0.63} &
  40.16\pmstd{2.20} &
  39.79\pmstd{1.70} \\
  \bottomrule
  \end{tabular}
\end{table}

Appendix Table~\ref{tab:ablation-small} reports architecture ablations of MoREGT-small (30M parameters), trained from scratch on the nine RelBench tasks of \texttt{rel-f1}, \texttt{rel-event} and \texttt{rel-trial}. The full model has the best mean score on both classification (79.14\% AUROC, 0.71 points above the next-best variant) and regression (44.40\% nMAE, 1.89 points below the next-best variant), and it is best on five of the nine tasks. The regression margin does not rest on the high-variance \texttt{site-success} task: without it, the full model still has the lowest mean nMAE (31.55\%, against 32.48\% for the next-best variant). The differences are clearest where variance across seeds is low. On \texttt{study-adverse}, every ablation raises nMAE by at least 1.93 points, and on \texttt{driver-top3} every ablation lowers AUROC by at least 1.52 points.

\begin{figure}[t]
    \centering
    \includegraphics[width=\figw{0.92\textwidth}{\columnwidth}]{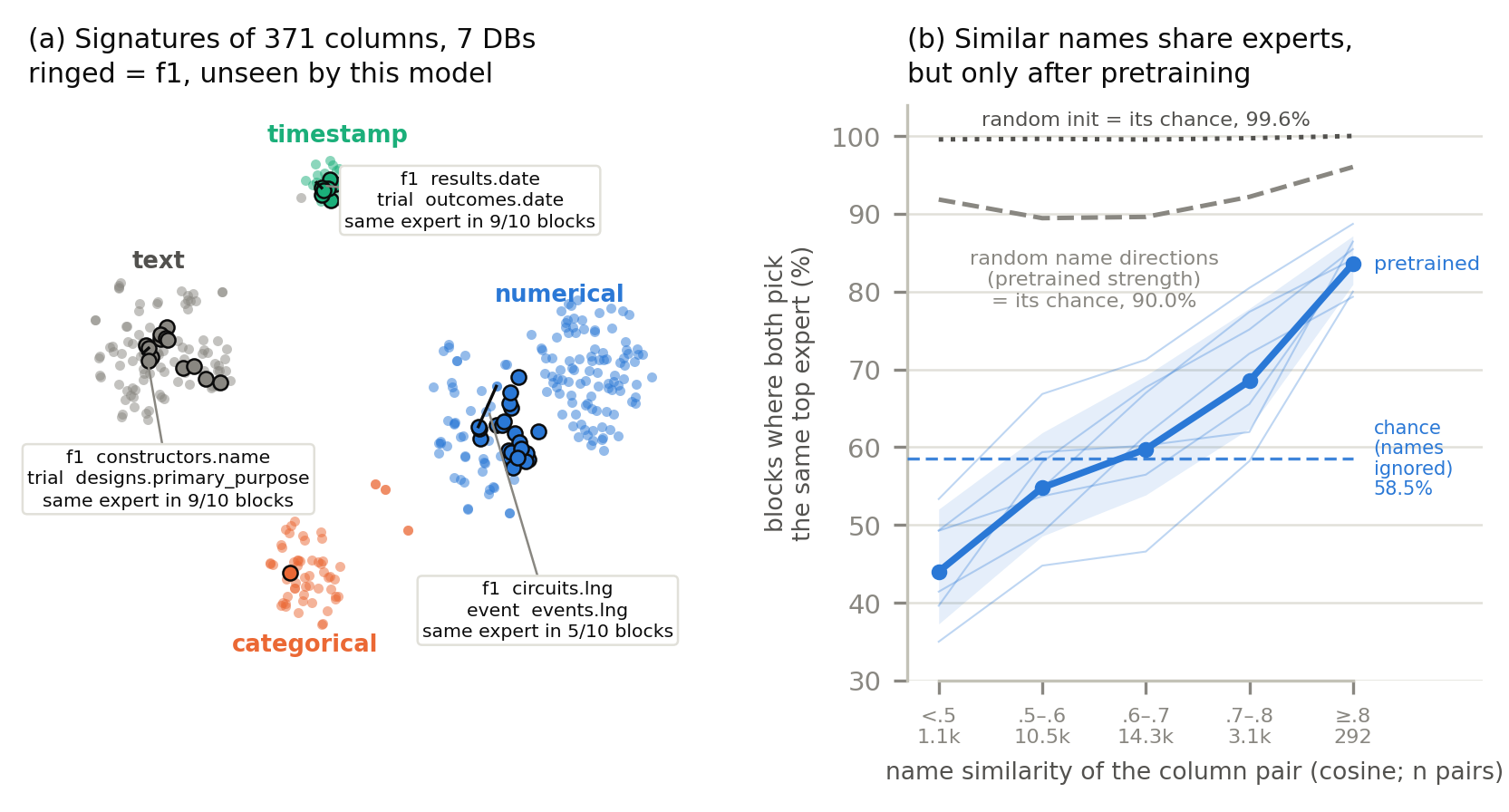}
    \caption{\textbf{Column names steer expert routing, and only after pretraining.} (a)~t-SNE of the routing signatures of all 371 columns under the trunk pretrained without rel-f1, coloured by modality. Callouts count the blocks (of 10) in which a rel-f1 column and its most similarly named same-modality column elsewhere pick the same top expert. (b)~Share of blocks in which a held-out column and a seen column of the same modality pick the same top expert, against the cosine similarity of their names, pooled over the seven LODO trunks (thin lines: folds; band: 95\% CI).}
    \label{fig:signature-routing}
\end{figure}

\noindent\textbf{Column names steer routing.} Since the router reads only the value-free signature, we route every column offline from the weights of each leave-one-database-out trunk and compare held-out columns with same-modality columns in the seen databases. Agreement is measured against chance: the rate any same-modality pair would show if names were ignored, which depends on how concentrated expert use is. Both controls sit at their own chance level near 100\%. At initialisation the modality embedding is about 150 times larger than the variation between names, so every column of a modality routes to the same expert. With the trained name projection's strength but random directions, names barely register: column-name embeddings occupy a narrow subspace (about 37 effective dimensions of 2,560), which random directions mostly miss. After pretraining, names move the signature as much as modality does. Chance falls to 58.5\%, and agreement rises with name similarity, from 44\% (cosine $<0.5$, below chance) to 84\% ($\geq 0.8$). For example, the unseen rel-f1 column \texttt{results.date} routes like rel-trial's \texttt{outcomes.date} in 9 of 10 blocks. Per fold, the most similarly named column shares the top expert 8.9 points more often than a random same-type column ($-0.1$ at initialisation, $+0.5$ with random directions), on six of seven folds. The exception, rel-event, has 100 of 122 columns named \texttt{c\_1}\ldots\texttt{c\_100}.

\noindent\textbf{Other feature learning methods.} Table~\ref{tab:featuretools-board} compares agent-derived columns against generic Featuretools transformations and evaluates their transfer to the RT backbone. On nine tasks, MoREGT with agent-derived columns outperforms Featuretools on eight, improving mean AUROC from 82.03 to 83.63 and reducing nMAE from 40.61 to 38.35. The largest differences are on user-attendance (8.81 nMAE) and site-success (5.16 nMAE), while on driver-dnf and study-adverse the gap is within one standard deviation. The exception is driver-position, where Featuretools reaches 32.56 nMAE, the lowest error of any method on that task. Applying the same agent-derived columns to RT yields substantially worse results (76.16 AUROC, 49.70 nMAE), even under the same fine-tuning recipe. These results suggest that the utility of derived columns can vary across relational architectures and fine-tuning configurations.
\begin{table}[t]
  \centering
  \caption{Agent-derived columns versus generic feature synthesis
  and the RT backbone. The MoREGT variants share the same LODO
  pretrained trunk; RT uses its own LODO trunk. All methods use
  the same fine-tuning recipe. TEST AUROC ($\uparrow$) and
  nMAE ($\downarrow$), in \%. MoREGT results are mean $\pm$
  std over 3 seeds.}
  \label{tab:featuretools-board}
  \footnotesize
  \setlength{\tabcolsep}{2.5pt}
  \begin{tabular}{@{}llccc@{}}
  \toprule
  \textbf{Dataset} & \textbf{Task} &
  \textbf{\begin{tabular}[b]{@{}c@{}}MoREGT\\+ agent\end{tabular}} &
  \textbf{\begin{tabular}[b]{@{}c@{}}MoREGT\\+ Featuretools\end{tabular}} &
  \textbf{\begin{tabular}[b]{@{}c@{}}RT\\+ agent\end{tabular}} \\
  \midrule
  \texttt{f1} & driver-dnf $\uparrow$ &
  \textbf{84.11}\pmstd{0.39} & 83.96\pmstd{0.76} & 82.93 \\
  \texttt{f1} & driver-top3 $\uparrow$ &
  \textbf{92.06}\pmstd{0.76} & 91.12\pmstd{0.75} & 85.52 \\
  \texttt{event} & user-ignore $\uparrow$ &
  \textbf{86.31}\pmstd{3.38} & 84.26\pmstd{1.34} & 71.88 \\
  \texttt{event} & user-repeat $\uparrow$ &
  \textbf{82.63}\pmstd{2.42} & 80.94\pmstd{0.72} & 73.56 \\
  \texttt{trial} & study-outcome $\uparrow$ &
  \textbf{73.05}\pmstd{1.72} & 69.89\pmstd{0.50} & 66.89 \\
  \midrule
  \texttt{f1} & driver-position $\downarrow$ &
  37.70\pmstd{1.98} & \textbf{32.56}\pmstd{1.05} & 39.76 \\
  \texttt{event} & user-attendance $\downarrow$ &
  \textbf{31.09}\pmstd{0.62} & 39.90\pmstd{1.90} & 56.42 \\
  \texttt{trial} & study-adverse $\downarrow$ &
  \textbf{9.75}\pmstd{0.27} & 9.97\pmstd{0.88} & 14.67 \\
  \texttt{trial} & site-success $\downarrow$ &
  \textbf{74.84}\pmstd{5.56} & 80.00\pmstd{7.31} & 87.95 \\
  \midrule
  \multicolumn{3}{@{}l}{\textit{Mean AUROC $\uparrow$}}
  \textbf{83.63} & 82.03 & 76.16 \\
  \multicolumn{3}{@{}l}{\textit{Mean nMAE $\downarrow$}}
  \textbf{38.35} & 40.61 & 49.70 \\
  \bottomrule
  \end{tabular}
\end{table}

\section{Conclusion}\label{sec:conclusion}
We presented MoREGT, a graph transformer with row and cell tokens, time- and role-aware attention and value-free expert routing, paired with a label-blind agent that adds validated single-row columns to the target database. Under the same LODO pretraining as RT, MoREGT improves on RT, and with derived columns it reaches the highest mean AUROC (79.63) and lowest nMAE (26.98) among the compared methods, despite pretraining on substantially smaller data. After pretraining, columns with similar names share experts across databases. This demonstrates that better architectures and target-specific features thus offer a route to stronger relational foundation for transfer across Web platforms, alongside larger pretraining corpora.

\bibliographystyle{ACM-Reference-Format}
\bibliography{sample-base}

\clearpage
\appendix
\renewcommand{\thefigure}{A\arabic{figure}}
\renewcommand{\thetable}{A\arabic{table}}
\renewcommand{\theHfigure}{A\arabic{figure}}
\renewcommand{\theHtable}{A\arabic{table}}
\setcounter{figure}{0}
\setcounter{table}{0}

\section{Implementation Details}\label{app:impl}

Table~\ref{tab:hparams} lists the value of every symbol introduced in Section~\ref{sec:method}. All seven leave-one-database-out trunks use the same values.

\begin{table*}[htbp]
  \centering
  \caption{Hyperparameters of MoREGT and of the agentic loop.}
  \label{tab:hparams}
  \small
  \begin{adjustbox}{max width=\textwidth}
  \begin{tabular}{@{}llr@{\qquad}llr@{}}
    \toprule
    & \textbf{Symbol} & \textbf{Value} & & \textbf{Symbol} & \textbf{Value} \\
    \midrule
    \multicolumn{3}{@{}l@{\qquad}}{\textit{Sampling and tokens}} & \multicolumn{3}{l@{}}{\textit{Experts}} \\
    Hops, neighbors per edge type & $N_{\mathrm{hop}}$, $N_{\mathrm{nbr}}$ & 2, 12 & Cell experts: routed, shared, top-$k$ & $N_{\mathrm{r}}$, $N_{\mathrm{s}}$, $k_{\mathrm{r}}$ & 8, 2, 2 \\
    Cells per example & $N_{\mathrm{cell}}$ & 1024 & Row experts: routed, shared & $N^{\mathrm{row}}_{\mathrm{r}}$, $N^{\mathrm{row}}_{\mathrm{s}}$ & 4, 1 \\
    Standardization constant & $\epsilon_\sigma$ & $10^{-6}$ & Expert hidden width & $d_{\mathrm{ff}}$ & 1024 \\
    Name-embedding dimension & $d_n$ & 2560 & Parameters (share in experts) & -- & 133.9M (95\%) \\
    \addlinespace[2pt]
    \multicolumn{3}{@{}l@{\qquad}}{\textit{Backbone}} & \multicolumn{3}{l@{}}{\textit{Agentic loop}} \\
    Width, blocks, attention heads & $d$, $L$, -- & 256, 10, 8 & Turn budget, feedback rounds & $N_{\mathrm{turn}}$, $R$ & 200, 3 \\
    Rotary frequencies & $K$, $[\omega_{\min},\omega_{\max}]$ & 8, $[0.3,5]$ & Range fraction, missing-rate slack & $q$, $\eta$ & 0.9, 5 points \\
    Signature width & $d_s$ & 128 & Correlation threshold & $r_{\max}$ & 0.999 \\
    Pooling heads, head width & $M$, $d_p$ & 4, 64 & & & \\
    \addlinespace[2pt]
    \multicolumn{3}{@{}l@{\qquad}}{\textit{Pretraining objective}} & & & \\
    Extra seed-row masking rate & $p_{\mathrm{mask}}$ & 0.5 & & & \\
    Loss weights & $\lambda_{\mathrm{cat}}$, $\lambda_{\mathrm{aux}}$, $\lambda_{\mathrm{bal}}$ & 4, 0.5, 0.01 & & & \\
    \bottomrule
  \end{tabular}
  \end{adjustbox}
\end{table*}
\noindent\textbf{Encoders.}
Each semantic type has one value encoder, shared by all its columns. A numerical value is standardized to $\tilde x_{i,j}=(x_{i,j}-\mu_j)/(\sigma_j+\epsilon_\sigma)$, where $\mu_j$ and $\sigma_j$ are the mean and standard deviation of column $j$, and encoded as $\mathbf{v}_{i,j}=\mathbf{w}_{\mathrm{num}}\tilde x_{i,j}+\mathbf{b}_{\mathrm{num}}$, with missing values mapped to $\tilde x_{i,j}=0$. A categorical value $c$ is encoded through its frozen embedding $\mathbf{n}_{j,c}$, free text through a frozen sentence encoder, and timestamps through a standardized year and sine--cosine harmonics of calendar fields. Column names, table names (as the string ``table $T$''), category values and FK roles are embedded once, offline, with Qwen3-Embedding-4B, and free-text cell values with MiniLM-L12-v2. No language model runs inside the training loop.

\noindent\textbf{Initial row embeddings and signatures.}
The initial embedding of row $i$, which the first block reads, is a bias-free projection of the mean of its cells' initial embeddings together with its signature
$\mathbf{U}^{(0)}_i=\mathbf{W}_u\big[\bar{\mathbf{H}}^{(0)}_i;\mathbf{S}_i\big],$
where $\bar{\mathbf{H}}^{(0)}_i$ is the mean of $\mathbf{H}^{(0)}_{i,j}$ over the cells of row $i$ that survive the cap $N_{\mathrm{cell}}$. For a row whose cells were all removed by the cap, $\bar{\mathbf{H}}^{(0)}_i=\mathbf{0}$, so its initial embedding depends only on its signature. The projections $\mathbf{W}_z$, $\mathbf{W}_\tau$, $\mathbf{W}_T$, $\mathbf{W}_\rho$ and $\mathbf{W}'_\tau$ of Eqs.~\eqref{eq:cell-sig} and~\eqref{eq:row-sig} are bias-free linear maps, and $\mathbf{E}_{\mathrm{st}}$ and $\mathbf{E}_{\mathrm{hop}}$ are lookup tables. The role $\rho_0(i)$ is found by a breadth-first search over the sampled row graph from the seed, with ties broken by the lowest-indexed frontier predecessor. Roles are direction-folded, so rows reached through the same FK from the parent side or from the child side share one role. The seed row takes a reserved role whose embedding is zero, so its signature is the RMSNorm of its table-name, hop-$0$ and age terms.

\noindent\textbf{Time in attention.}
Ages are computed in float64 with $\Delta_i$ in seconds on every database. In each head of width $d_h$, half of the dimensions are rotated, as $K=d_h/4$ planar pairs with one frequency each, and the other half are left unrotated. The rotation angle of row $i$ is $\omega_f\tau_i$ for a timed row and $0$ for an untimed row. An unrotated row would therefore look as recent as a row with $\Delta_i=0$, which the untimed offset corrects. The clamped indicator of row $i$ is computed from the raw times as $t_i>t^*$ and is set only for timed rows; it is also the last entry of $\boldsymbol{\varphi}_i$. Two learned scalars, $b_{\mathrm{untimed}}$ and $b_{\mathrm{clamped}}$, both initialized to zero, are added to the attention score of a pair when the query row or the key row is untimed or clamped, respectively. Both are shared by all heads and by cell and row attention. $\mathrm{CellAttn}$ receives only a padding mask, the log-ages and these two indicators, so it has no structural mask and no row identifier.

\noindent\textbf{Pooling.}
$\mathrm{Pool}$ is cross-attention with $M$ heads of width $d_p$ from each row to its own cells. Head $m$ weights the cells of row $i$ by $\alpha^m_{i,j}$, and the heads' outputs are concatenated into the update that the third line of Eq.~\eqref{eq:block} adds to the row embedding:
\begin{equation}\label{eq:pool}
\begin{aligned}
\alpha^{m}_{i,j}&=\operatorname*{softmax}_{j\in\mathcal{J}(i)}\frac{\big(\mathbf{W}^{Q}_{m}[\mathbf{U}_i;\mathbf{S}_i]\big)^{\top}\mathbf{W}^{K}\mathbf{n}_{j}}{\sqrt{d_p}},\\
\mathrm{Pool}_i&=\mathbf{W}^{O}\,\big\Vert_{m=1}^{M}\textstyle\sum_{j\in\mathcal{J}(i)}\alpha^{m}_{i,j}\,\mathbf{W}^{V}\mathbf{H}''_{i,j},
\end{aligned}
\end{equation}
where $\mathbf{U}_i=\mathbf{U}^{(\ell-1)}_i$, $\mathcal{J}(i)$ is the set of columns of row $i$ whose cells survive the cap, $\mathbf{W}^{Q}_{m}$ projects $[\mathbf{U}_i;\mathbf{S}_i]$ to width $d_p$, $\mathbf{W}^{K}$ and $\mathbf{W}^{V}$ project column-name embeddings and cell embeddings to width $d_p$ and are shared by all heads, $\Vert$ concatenates the $M$ head outputs, and $\mathbf{W}^{O}\in\mathbb{R}^{d\times Md_p}$. All four projections are bias-free. A row whose cells were all removed by the cap receives $\mathrm{Pool}_i=\mathbf{0}$.

\noindent\textbf{Experts and routing regularizers.}
Every expert, routed or shared, is a bias-free SwiGLU network $\mathrm{E}(\hat{\mathbf{H}}_{i,j})=\mathbf{W}_3\big(\operatorname{SiLU}(\mathbf{W}_1\hat{\mathbf{H}}_{i,j})\odot\mathbf{W}_2\hat{\mathbf{H}}_{i,j}\big)$ of hidden width $d_{\mathrm{ff}}$. The cell router $\mathbf{W}_g$ of Eq.~\eqref{eq:cell-moe} is a linear map on $\mathbf{Z}_{i,j}$. The row router is a cosine router~\citep{chi2022representation} on the row signature, which keeps $k_{\mathrm{r}}$ of the $N^{\mathrm{row}}_{\mathrm{r}}$ routed row experts with gates
\begin{equation}\label{eq:row-router}
\mathbf{g}^{\mathrm{row}}_i=\operatorname{softmax}\Big(\operatorname{TopK}_{k_{\mathrm{r}}}\Big(\big[\cos(\mathbf{S}_i,\mathbf{w}^{\mathrm{row}}_e)/\kappa\big]_{e=1}^{N^{\mathrm{row}}_{\mathrm{r}}}\Big)\Big),
\end{equation}
where $\mathbf{w}^{\mathrm{row}}_e$ is the router row of row expert $e$, $\cos$ is cosine similarity, and $\kappa=\exp(\nu)$ is a temperature with $\nu$ a learned scalar initialized at zero. Every row also passes through the $N^{\mathrm{row}}_{\mathrm{s}}$ shared row experts.

Column types are long-tailed, and since all cells of a type share a route, a uniform load-balancing loss~\citep{fedus2022switch} could equalize load only by merging types. We instead keep the routing directions of the cell router distinct without constraining usage:
\begin{equation}\label{eq:ortho}
\Omega_{\mathrm{cell}}=\big\|\hat{\mathbf{W}}_g\hat{\mathbf{W}}_g^{\top}-\mathbf{I}\big\|_F^2=\textstyle\sum_{e\neq e'}\cos^2(\mathbf{w}_e,\mathbf{w}_{e'}),
\end{equation}
where $\mathbf{w}_e$ is row $e$ of $\mathbf{W}_g$, $\hat{\mathbf{W}}_g$ is $\mathbf{W}_g$ with $\ell_2$-normalized rows, $\mathbf{I}$ is the $N_{\mathrm{r}}\times N_{\mathrm{r}}$ identity, and the sum runs over ordered pairs of distinct routed experts. $\Omega_{\mathrm{row}}$ is the same penalty on the row-router rows $\mathbf{w}^{\mathrm{row}}_e$. The row router also receives a Switch-style load-balancing term~\citep{fedus2022switch}, computed over the rows that are not padding:
\begin{equation}\label{eq:balance}
B_{\mathrm{row}}=N^{\mathrm{row}}_{\mathrm{r}}\textstyle\sum_{e=1}^{N^{\mathrm{row}}_{\mathrm{r}}}f_e\,\bar g_e,
\end{equation}
where $f_e$ is the fraction of these rows whose highest-scoring expert is $e$, and $\bar g_e$ is the mean over these rows of the gate of $e$ in Eq.~\eqref{eq:row-router}, which is zero for rows that do not keep $e$. The routing term of the loss sums both levels over the blocks,
\begin{equation}\label{eq:aux}
\mathcal{L}_{\mathrm{aux}}=\textstyle\sum_{\ell=1}^{L}\big(\Omega^{(\ell)}_{\mathrm{cell}}+\lambda_{\mathrm{aux}}\Omega^{(\ell)}_{\mathrm{row}}+\lambda_{\mathrm{bal}}B^{(\ell)}_{\mathrm{row}}\big),
\end{equation}
where the superscript $(\ell)$ marks the terms of block $\ell$. Pretraining and fine-tuning both add $\lambda_{\mathrm{aux}}\mathcal{L}_{\mathrm{aux}}$ to the task loss, so the three terms carry effective weights $\lambda_{\mathrm{aux}}$, $\lambda_{\mathrm{aux}}^2$ and $\lambda_{\mathrm{aux}}\lambda_{\mathrm{bal}}$, that is, $0.5$, $0.25$ and $0.005$.

  \begin{table*}[t]
  \centering
  \caption{Columns derived by the agent for MoREGT (with agent), with one example per task: 210 of 230 proposed columns kept after screening (\emph{Acc./prop.}), \$28.59 of API cost in total. The full list, with every column's expression, units and rationale, is in the repository (\texttt{results/derive/accepted\_columns.md}).}
  \label{tab:agent-columns}
  \footnotesize
  \setlength{\tabcolsep}{4pt}
  \begin{tabularx}{\textwidth}{@{}l l c >{\raggedright\arraybackslash}X@{}}
  \toprule
  \textbf{Dataset} & \textbf{Task} & \textbf{Acc./prop.} & \textbf{Example column} \\
  \midrule
  \texttt{amazon} & item-churn & 9/9 & \texttt{review.verified\_\allowbreak{}complaint} $=$ \texttt{(rating <= 2) * verified} \\
   & item-ltv & 11/11 & \texttt{product.charm\_\allowbreak{}priced} $=$ \texttt{(price \% 1) >= 0.9} \\
   & user-churn & 9/9 & \texttt{review.verified\_\allowbreak{}disappointment} $=$ \texttt{(5 - rating) * verified} \\
   & user-ltv & 9/9 & \texttt{product.impulse\_\allowbreak{}priced\_\allowbreak{}item} $=$ \texttt{(price > 0) * (price < 5)} \\
  \addlinespace[2pt]
  \texttt{avito} & ad-ctr & 12/12 & \texttt{SearchStream.position\_\allowbreak{}discounted\_\allowbreak{}prior} $=$ \texttt{HistCTR / log1p(Position)} \\
   & user-clicks & 9/9 & \texttt{SearchStream.paid\_\allowbreak{}top\_\allowbreak{}slot\_\allowbreak{}impression} $=$ \texttt{(ObjectType == 3) * (Position <= 1)} \\
   & user-visits & 9/9 & \texttt{SearchInfo.search\_\allowbreak{}filters\_\allowbreak{}applied} $=$ \texttt{(CategoryID > 0) + (LocationID > 0)} \\
  \addlinespace[2pt]
  \texttt{event} & user-attendance & 9/9 & \texttt{event\_\allowbreak{}interest.invitation\_\allowbreak{}uptake} $=$ \texttt{invited * interested} \\
   & user-ignore & 9/9 & \texttt{event\_\allowbreak{}interest.ignored\_\allowbreak{}invitation} $=$ \texttt{invited * (1 - interested) * (1 - not\_\allowbreak{}interested)} \\
   & user-repeat & 9/9 & \texttt{event\_\allowbreak{}interest.organic\_\allowbreak{}interest} $=$ \texttt{interested * (1 - invited)} \\
  \addlinespace[2pt]
  \texttt{f1} & driver-dnf & 14/14 & \texttt{results.mechanical\_\allowbreak{}dnf} $=$ \texttt{(statusId > 4) * (statusId < 11)} \\
   & driver-position & 9/9 & \texttt{results.positions\_\allowbreak{}gained} $=$ \texttt{grid - positionOrder} \\
   & driver-top3 & 12/12 & \texttt{results.front\_\allowbreak{}row\_\allowbreak{}start} $=$ \texttt{(grid >= 1) * (grid <= 3)} \\
  \addlinespace[2pt]
  \texttt{hm} & item-sales & 9/9 & \texttt{transactions.online\_\allowbreak{}revenue} $=$ \texttt{price * (sales\_\allowbreak{}channel\_\allowbreak{}id - 1)} \\
   & user-churn & 9/10 & \texttt{customer.subscribed\_\allowbreak{}not\_\allowbreak{}active} $=$ \texttt{(FN > 0) * (1.0 - (Active > 0) * 1.0)} \\
  \addlinespace[2pt]
  \texttt{stack} & post-votes & 12/12 & \texttt{votes.score\_\allowbreak{}vote\_\allowbreak{}sign} $=$ \texttt{1 * (VoteTypeId == 2) - 1 * (VoteTypeId == 3)} \\
   & user-badge & 12/12 & \texttt{badges.achievement\_\allowbreak{}badge\_\allowbreak{}prestige} $=$ \texttt{(TagBased == 0) * (4 - Class)} \\
   & user-engagement & 11/11 & \texttt{posts.is\_\allowbreak{}answer} $=$ \texttt{1 * (PostTypeId == 2)} \\
  \addlinespace[2pt]
  \texttt{trial} & site-success & 11/19 & \texttt{studies.is\_\allowbreak{}underpowered\_\allowbreak{}comparative} $=$ \texttt{(number\_\allowbreak{}of\_\allowbreak{}arms >= 2) * (enrollment < 30 * number\_\allowbreak{}of\_\allowbreak{}arms)} \\
   & study-adverse & 8/11 & \texttt{studies.log\_\allowbreak{}enrollment\_\allowbreak{}per\_\allowbreak{}arm} $=$ \texttt{log1p(enrollment / maximum(number\_\allowbreak{}of\_\allowbreak{}arms, 1))} \\
   & study-outcome & 8/16 & \texttt{outcome\_\allowbreak{}analyses.effect\_\allowbreak{}signal\_\allowbreak{}fraction} $=$ \texttt{abs(param\_\allowbreak{}value) / (abs(param\_\allowbreak{}value) + abs(dispersion\_\allowbreak{}value) + 0.000001)} \\
  \bottomrule
  \end{tabularx}
  \end{table*}

\noindent\textbf{Pretraining.}
We train each trunk with AdamW ($\beta=(0.9,0.95)$, learning rate $3\times10^{-4}$, weight decay $0.1$ on matrices only), with $5\%$ warmup followed by linear decay over a fixed budget of 18 hours, gradient clipping at $1.0$ and bfloat16 precision. The batch is 64 examples per GPU on 4 H100 GPUs, which gives 208k--237k steps per trunk. Each step draws one seed table, with probability proportional to its number of maskable columns times its number of eligible rows, normalized so that each pretraining database receives an equal share of steps. Pretraining seeds lie before the RelBench validation timestamp, and the last $10\%$ of that window is held out to select the checkpoint.

\noindent\textbf{Fine-tuning and model selection.}
Fine-tuning uses AdamW with a constant learning rate of $3\times10^{-4}$, weight decay $0.01$ and batch size 128 (64 on \texttt{rel-event}, for memory), for at most 12 epochs of 200 batches, with early stopping after 4 epochs without improvement. We report the epoch with the best RelBench validation metric, averaged over three seeds.

\section{Additional Results}

Table~\ref{tab:agent-columns} lists the columns the agent added for MoREGT (with agent). Over the 21 tasks it proposed 230 columns, and screening kept 210, between 8 and 14 per task, for \$28.59 of API cost in total. All proposals were kept on 17 of the 21 tasks; the 20 rejections are on \texttt{hm} user-churn (1) and the three \texttt{trial} tasks (19). The example columns in the table are of three kinds: indicators that group codes or thresholds (\texttt{results.\allowbreak{}mechanical\_\allowbreak{}dnf}, \texttt{posts.\allowbreak{}is\_\allowbreak{}answer}), products of a flag with another flag or an amount (\texttt{review.\allowbreak{}verified\_\allowbreak{}complaint}, \texttt{transactions.\allowbreak{}online\_\allowbreak{}revenue}), and differences, ratios and logarithms (\texttt{results.\allowbreak{}positions\_\allowbreak{}gained}, \texttt{studies.\allowbreak{}log\_\allowbreak{}enrollment\_\allowbreak{}per\_\allowbreak{}arm}). Compared with MoREGT (w/o agent), the derived columns raise AUROC by 4.12 points on \texttt{trial} study-outcome, 3.10 on \texttt{avito} user-clicks and 2.69 on \texttt{hm} user-churn, and lower nMAE by 3.46 points on \texttt{trial} study-adverse and 2.80 on site-success (Tables~\ref{tab:board-clf} and~\ref{tab:board-reg}). On the \texttt{f1} tasks and on \texttt{event} user-repeat, they change scores by at most 0.21 points.

\begin{figure}[t]
  \centering
  \includegraphics[width=\figw{0.55\textwidth}{0.95\columnwidth}]{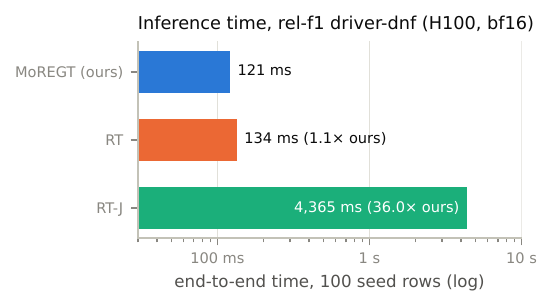}
  \caption{\textbf{Inference time.} 100 rel-f1 \texttt{driver-dnf} test seeds.}
  \label{fig:latency}
\end{figure}

\paragraph{Inference cost.}
  Let a seed's context hold $S$ cells in $R$ rows, and let the model have $L$ blocks of width $d$ and FFN width $d_{\mathrm{ff}}$. One MoREGT block costs
  \begin{equation}
    \adjustbox{max width=\linewidth}{$\displaystyle
    \underbrace{O(S^2 d)}_{\text{cell attention}}
    + \underbrace{O(R^2 d)}_{\text{row attention}}
    + \underbrace{O\!\big((k+n_s)\,S\,d\,d_{\mathrm{ff}} + (k_r+n_r)\,R\,d\,d_{\mathrm{ff}}\big)}_{\text{cell and row MoE FFNs}}
    + O(S d^2),$}
    \label{eq:cost}
  \end{equation}
  where the last term covers the projections, row pooling and broadcast. Experts are dispatched sparsely: each cell passes through its $k=2$ routed and $n_s=2$ shared experts only, so the FFN term does not grow with the number of experts ($M=8$). Row attention costs $(R/S)^2 \approx 3\%$ of cell attention. The forward pass is therefore $\Theta(L S^2 d)$, the same order as RT, whose every layer runs a full attention over all $S$ cells next to its column, feature and neighbour attentions. RT-J drops the full attention and sorts cells by column, so its column attention is block-diagonal and its block-sparse kernel skips most blocks, but its worst case is still $O(S^2 d)$. Both RT models also materialise $S \times S$ boolean masks, including an $S \times S \times f$ foreign-key comparison for $f$ foreign keys per cell; we build only a padding block mask with $(S/128)^2$ entries. At our width ($d=256$, $d_{\mathrm{ff}}=1024$, four active experts per cell), the linear FFN term in \eqref{eq:cost} exceeds the quadratic attention term up to $S \approx 6{,}000$ cells, so at deployed sizes our cost per seed grows almost linearly in $S$.
  Building the context is linear in its size for our temporal neighbour sampler and for RT's breadth-first sampler. RT-J additionally runs $W = 10^4$ random walks of length $\ell = 20$ per seed to find in-context examples, at $O(S + W\ell)$ per seed whatever $S$ is. As shown in Figure~\ref{fig:latency}, MoREGT is fastest for inferring 100 rel-f1 \texttt{driver-dnf} test seeds. MoREGT is about 36x faster than RT-J while achieving comparable performance.

Each variant in Table~\ref{tab:ablation-small} removes one component: Dense FFN replaces every mixture of experts with a dense feed-forward layer, Hidden routing routes on the hidden state instead of the value-free signature, No time attn.\ removes the log-age terms from attention, and No role bias removes the FK-role bias of Eq.~\eqref{eq:rolebias}. Removing time from attention gives the largest drop in both means (1.93 AUROC and 2.83 nMAE), followed in classification by removing the role bias (1.75 AUROC); in regression, Dense FFN is close behind (2.77 nMAE), and both regression gaps come mostly from \texttt{site-success}. The routing ablations matter most on \texttt{driver-dnf}, where Dense FFN and Hidden routing lose 2.30 and 2.46 AUROC with standard deviations below 0.6, against 0.65 for No time attn. On \texttt{study-outcome}, the attention components matter more than routing: removing the time terms or the role bias lowers AUROC by 1.97 and 2.33, against 0.39 and 1.02 for the routing variants, although these gaps are only one to two standard deviations. On \texttt{rel-event}, no variant differs from the full model by more than one standard deviation.

\begin{table*}[t]
  \centering
  \caption{Ablations of MoREGT-small (30M parameters) trained from scratch: TEST AUROC (\%, $\uparrow$) for classification and nMAE (\%, $\downarrow$) for regression. Mean $\pm$ std over 3 seeds. \textbf{Bold}: best per row.}
  \label{tab:ablation-small}
  \setlength{\tabcolsep}{3pt}
  \footnotesize
  \begin{adjustbox}{max width=\textwidth}
  \begin{tabular}{@{}ll ccccc@{}}
  \toprule
   & & \textbf{Full} & \textbf{Dense} & \textbf{Hidden} & \textbf{No time} & \textbf{No role} \\
   & & & \textbf{FFN} & \textbf{routing} & \textbf{attn.} & \textbf{bias} \\
  \midrule
  \multicolumn{7}{@{}l}{\textit{Classification (AUROC $\uparrow$)}} \\
  \texttt{f1}    & driver-dnf    & \textbf{84.12}\pmstd{0.13} & 81.82\pmstd{0.09} & 81.66\pmstd{0.54} & 83.47\pmstd{0.65} & 82.34\pmstd{0.22} \\
                 & driver-top3   & \textbf{93.04}\pmstd{0.73} & 91.03\pmstd{0.69} & 91.52\pmstd{0.25} & 90.97\pmstd{0.42} & 91.16\pmstd{0.37} \\
  \texttt{event} & user-repeat   & 76.88\pmstd{5.00} & 76.54\pmstd{1.84} & \textbf{77.08}\pmstd{0.52} & 72.85\pmstd{5.51} & 73.84\pmstd{5.96} \\
                 & user-ignore   & 76.07\pmstd{3.07} & \textbf{77.60}\pmstd{5.06} & 75.02\pmstd{1.61} & 75.15\pmstd{5.86} & 76.35\pmstd{5.65} \\
  \texttt{trial} & study-outcome & \textbf{65.57}\pmstd{1.29} & 65.18\pmstd{0.84} & 64.55\pmstd{1.73} & 63.60\pmstd{1.38} & 63.24\pmstd{1.40} \\
                 & \textit{Mean} & \textbf{79.14} & 78.43 & 77.97 & 77.21 & 77.39 \\
  \midrule
  \multicolumn{7}{@{}l}{\textit{Regression (nMAE $\downarrow$)}} \\
  \texttt{f1}    & driver-position & 38.24\pmstd{2.95} & \textbf{38.04}\pmstd{1.16} & 39.84\pmstd{1.29} & 39.32\pmstd{1.16} & 39.45\pmstd{0.20} \\
  \texttt{event} & user-attendance & 43.43\pmstd{6.56} & 45.36\pmstd{5.08} & 43.34\pmstd{3.96} & \textbf{43.06}\pmstd{2.19} & 46.29\pmstd{6.23} \\
  \texttt{trial} & study-adverse   & \textbf{12.99}\pmstd{0.28} & 14.92\pmstd{0.04} & 15.02\pmstd{0.10} & 15.06\pmstd{0.24} & 15.08\pmstd{0.44} \\
                 & site-success    & \textbf{82.93}\pmstd{5.53} & 90.37\pmstd{0.78} & 86.94\pmstd{5.84} & 91.47\pmstd{6.25} & 85.53\pmstd{5.43} \\
                 & \textit{Mean}   & \textbf{44.40} & 47.17 & 46.29 & 47.23 & 46.59 \\
  \bottomrule
  \end{tabular}
  \end{adjustbox}
\end{table*}

\end{document}